\documentclass[sigplan,10pt]{acmart}

\renewcommand\footnotetextcopyrightpermission[1]{}

\AtBeginDocument{%
  \providecommand\BibTeX{{%
    \normalfont B\kern-0.5em{\scshape i\kern-0.25em b}\kern-0.8em\TeX}}}

\usepackage{xcolor}
\usepackage{soul}
\usepackage{multirow}
\usepackage{makecell}
\sethlcolor{red!15}

\newcommand{\sys}{MosaicKV}

\begin{document}

\title{{\sys}: Serving Long-Context LLM with Dynamic Two-D KV Cache Compression}


\author{Sheng Qiang}
\affiliation{%
  \institution{Institute of Parallel and Distributed Systems, Shanghai Jiao Tong University}
  \city{Shanghai}
  \country{China}
}
\email{qiangsheng@sjtu.edu.cn}

\author{Ruiwei Chen}
\affiliation{%
  \institution{Institute of Parallel and Distributed Systems, Shanghai Jiao Tong University}
  \city{Shanghai}
  \country{China}
}
\email{chenruiwei@sjtu.edu.cn}

\author{Yinpeng Wu}
\affiliation{%
  \institution{Institute of Parallel and Distributed Systems, Shanghai Jiao Tong University}
  \city{Shanghai}
  \country{China}
}
\email{wyp1536481268@foxmail.com}

\author{Jinyu Gu}
\affiliation{%
  \institution{Institute of Parallel and Distributed Systems, Shanghai Jiao Tong University}
  \city{Shanghai}
  \country{China}
}
\email{gujinyu@sjtu.edu.cn}

\author{Zhichao Hua}
\authornote{Corresponding author.}
\affiliation{%
  \institution{Institute of Parallel and Distributed Systems, Shanghai Jiao Tong University}
  \city{Shanghai}
  \country{China}
}
\email{zchua@sjtu.edu.cn}

\author{Yubin Xia}
\affiliation{%
  \institution{Institute of Parallel and Distributed Systems, Shanghai Jiao Tong University}
  \city{Shanghai}
  \country{China}
}
\email{xiayubin@sjtu.edu.cn}

\author{Binyu Zang}
\affiliation{%
  \institution{Institute of Parallel and Distributed Systems, Shanghai Jiao Tong University}
  \city{Shanghai}
  \country{China}
}
\email{byzang@sjtu.edu.cn}

\author{Haibo Chen}
\affiliation{%
  \institution{Institute of Parallel and Distributed Systems, Shanghai Jiao Tong University}
  \city{Shanghai}
  \country{China}
}
\email{haibochen@sjtu.edu.cn}

\begin{abstract}
Long-context LLM services now sustain prompts with hundreds of thousands to millions of tokens, making the key-value (KV) cache a first-order serving cost.
Because the cache grows linearly with context length, it can exhaust GPU memory, force smaller batches, and reduce serving throughput.
Prior KV cache compression techniques typically target only the sequence dimension or only the channel dimension, which leaves limited headroom as context windows scale.
Compressing both dimensions promises higher memory reduction, but applying the two forms of compression directly leads to significant accuracy loss.

This paper introduces {\sys}, a dynamic two-D (dimensional) KV cache compression system for extremely long-context serving.
{\sys} uses \emph{dynamic two-D compression} to address the accuracy challenge, exploiting the non-uniform importance distribution of elements within the KV cache.
Instead of applying one compression pattern globally, {\sys} identifies important elements for each KV vector and selects compression strategies at the granularity of KV cache segments.
To address the performance challenge, where fine-grained sparsity and compression management overhead can offset the gains from compression, {\sys} introduces \emph{compressed KV cache management}.
This mechanism uses underutilized GPU and CPU resources to maintain compressed KV caches and accelerate attention computation.
%
Evaluation on an H800 GPU with multiple LLMs shows that {\sys} delivers up to \textbf{16$\times$} attention speedup, \textbf{4.8$\times$} lower decode latency, and \textbf{7.3$\times$} higher throughput than the uncompressed baseline.
At the same time, it reduces memory usage by \textbf{3$\times$} and incurs only \textbf{1.76\%} average accuracy loss on LongBench and RULER.
\end{abstract}

\maketitle

\pagestyle{plain}

\section{Introduction}
\label{sec:intro}

Large language models (LLMs) are increasingly deployed in scenarios that demand extremely long contexts, such as repository-level code generation~\cite{zhang_repocoder_2023,zhang_codeagent_2024}, multi-document question answering~\cite{wang_knowledge_2024,cheng_dualrag_2025}, and multi-turn agentic workflows~\cite{li_beyond_2026,wei_webagent-r1_2025,claudecode,openclaw}.
In response, frontier models~\cite{noauthor_llms_2026,noauthor_introducing_gpt5.4_nodate,noauthor_gemini3_2025,anthropic_whats_new_claude_46_2026,noauthor_LLaMA4_2025} have rapidly extended their supported context windows to 1M tokens, e.g., GPT-5.4~\cite{noauthor_introducing_gpt5.4_nodate} and Gemini~3.1 Pro~\cite{noauthor_gemini3_2025}.

However, long contexts impose a severe memory bottleneck through the key-value (KV) cache, whose size scales linearly with the sequence length during inference.
For instance, serving 8 concurrent LLaMA-3.1-8B requests at 128K tokens requires 128\,GB for the KV cache alone---already 8$\times$ the model weights; at 1M tokens, this grows to 1\,TB, 64$\times$ the model weights.
Such memory pressure exhausts GPU capacity, forces drastic reductions in batch size, and significantly increases per-step decode latency, ultimately degrading serving throughput.

Prior work addresses this bottleneck through KV cache compression~\cite{zhang_h2o_2023,tang_quest_2024,xu_think_2024,lee_infinigen_2024,sun_shadowkv_2025}, which operates along two dimensions: \emph{sequence compression} and \emph{channel compression}.
Sequence compression retains only a subset of historical tokens for attention computation.
For example, Quest~\cite{tang_quest_2024} performs query-aware token selection at runtime to achieve both high accuracy and low latency.
Channel compression, in contrast, retains a subset of channels from each token's KV vectors.
For example, ThinK~\cite{xu_think_2024} computes a uniform channel mask from the prefill key cache and applies it to select important elements across all key vectors.
However, each dimension alone cannot benefit from the other, making it difficult to further increase compression rates as context windows continue to expand.
A natural next step is therefore to combine the two compression dimensions to achieve higher compression rates for extremely long-context serving.


\textbf{Challenge~1: Accuracy Loss.}
Recent works attempt to combine the two dimensions of compression.
For example, ThinK~\cite{xu_think_2024} first applies channel compression to key vectors and then selects important tokens for attention computation.
However, it observes that accuracy degrades significantly when both channel and sequence compression are applied to the KV cache simultaneously.
Consequently, ThinK applies two-dimensional compression only to key vectors, while value vectors undergo sequence compression alone.
Other existing works, including InfiniGen~\cite{lee_infinigen_2024} and ShadowKV~\cite{sun_shadowkv_2025}, also cannot apply two-dimensional compression to both key and value vectors throughout token selection and attention computation.
The fundamental reason is that na\"{i}vely applying such full-lifecycle two-dimensional compression significantly degrades model accuracy.
In our evaluation, applying a 70\% channel compression rate on top of sequence compression causes an 82.8\% accuracy loss (Figure~\ref{fig:quest_svd_acc}).

\begin{figure*}[htb]
  \centering
  \includegraphics[width=\linewidth]{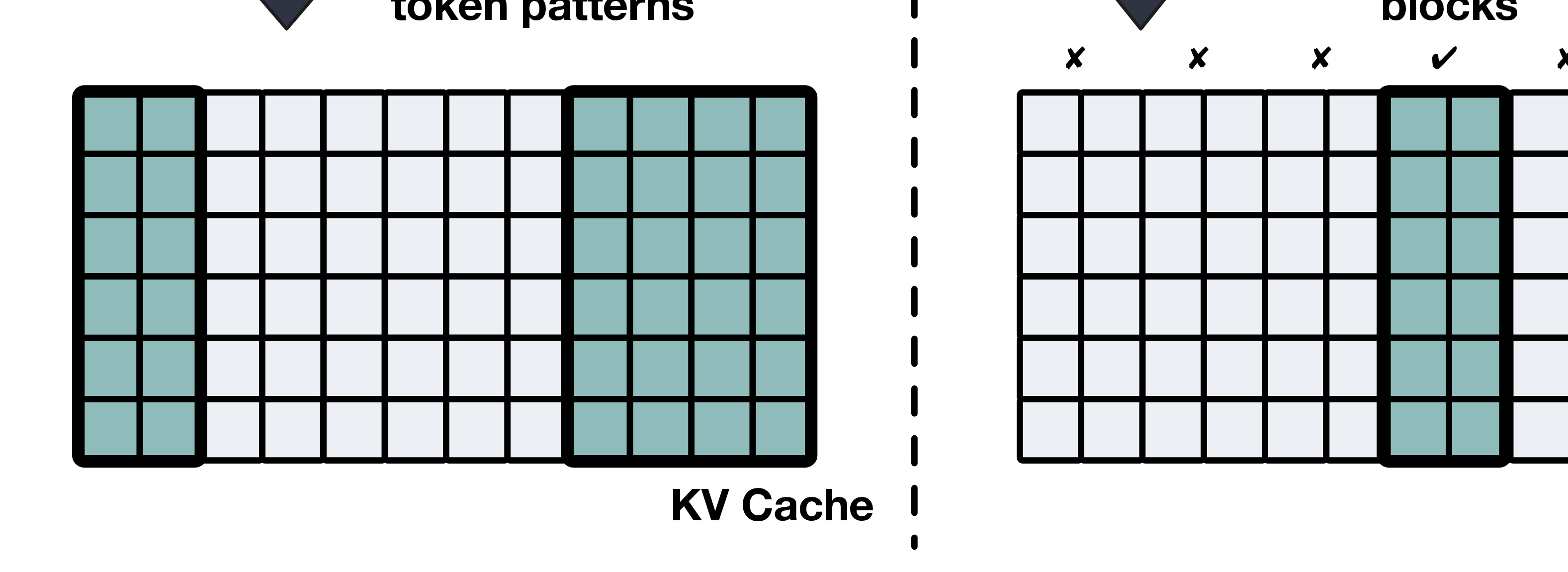}
  \caption{Different KV cache compression methods.}
  \label{fig:related}
\end{figure*}

\emph{Observation \& Approach~1: Dynamic Two-D Compression.}
We observe that the importance distribution of elements is non-uniform within the KV cache.
Existing works do not account for this variation and apply a one-size-fits-all compression strategy to the entire KV cache, leading to significant accuracy degradation under two-dimensional compression.
To address these limitations, this paper introduces {\sys}, a novel framework that exploits the differentiated importance distribution of the KV cache to serve long-context LLM inference with low memory usage, high decode performance, and low accuracy loss.
To overcome the accuracy challenge, {\sys} introduces \textbf{dynamic two-D (dimensional) compression}, which performs two-dimensional compression with a dynamic compression strategy.

{\sys} exploits two levels of differentiation.
First, the importance distribution differs across individual KV vectors, even after SVD rotation, as shown in Figure~\ref{fig:kv_cache_visualization}.
Only 62.28\% of the top-25\% important elements reside in the globally top-25\% important channels, even with a small 256-token context.
Unlike existing works that apply a uniform channel mask to all KV vectors, {\sys} introduces \emph{per-vector element selection}, which identifies the important elements for each KV vector individually.
Second, the element distribution characteristics vary across KV cache regions.
For example, in some regions, element magnitudes exhibit high variance with pronounced outliers, while in others, elements are distributed more uniformly with similar magnitudes.
Based on this observation, {\sys} introduces a \emph{dynamic compression strategy} that partitions the KV cache into multiple segments and applies a suitable strategy to each segment.
The dynamic two-D compression method enables {\sys} to apply two-dimensional KV cache compression throughout the entire decode stage, increasing the compression rate while preserving model accuracy.

\textbf{Challenge~2: Performance Degradation.}
Although {\sys}'s dynamic two-D compression achieves a high KV cache compression rate, it introduces fine-grained sparse memory access patterns and substantial memory management overhead.
Both of them hurt inference performance, potentially even worse than the uncompressed baseline.
For the former, per-vector element selection produces a fine-grained sparse KV cache matrix with poor cache locality.
Accessing such a matrix significantly increases the GPU cache miss rate, negating the potential speedup from compression.
For the latter, the dynamic compression strategy introduces heavy KV cache management overhead, including memory partitioning, SVD computation, and recompression, into the critical decode stage, further degrading performance.

\emph{Observation \& Approach~2: Compressed KV Cache Management.}
We observe that attention computation dominates both memory usage and latency during long-context decoding.
Moreover, the decode-stage attention exhibits a highly imbalanced workload: GPU memory bandwidth is fully saturated, while the GPU compute units (especially CUDA cores) and the CPU remain significantly underutilized, even with FlashAttention~\cite{dao2022flashattention}.
Profiling FlashInfer~\cite{ye2025flashinfer} during the decode stage of a 256K-token request on LLaMA-3.1-8B with an A800 GPU, we find that memory bandwidth utilization reaches 90.5\%, while CUDA core utilization is only 10.35\% and almost all CPU cores are idle.
To overcome the performance challenge, {\sys} introduces a novel \textbf{compressed KV cache management} method that exploits these underutilized resources to manage the compressed KV cache and accelerate attention computation.

{\sys} first introduces \emph{packed sparse attention} to accelerate attention computation over the two-D compressed KV cache.
It packs the sparse KV cache matrix into a dense format in GPU memory and leverages the underutilized CUDA cores, instead of tensor cores, to perform attention computation without unpacking.
This approach significantly reduces both the GPU cache miss rate and GPU memory usage, thereby accelerating attention computation.

Furthermore, {\sys} introduces \emph{heterogeneous double compression buffering} to reduce the overhead of compressed KV cache management.
A GPU-side buffer stores temporarily compressed, newly generated KV vectors that are used for decode computation in the interim.
A CPU-side buffer is used to perform asynchronous KV cache management operations, including partitioning KV cache segments, computing SVD rotation matrices, generating compression strategies, and recompressing new segments.
A switching mechanism replaces the temporarily compressed KV segment with the final compressed version without blocking the decode stage.


We implement a prototype of {\sys} and evaluate it on an H800 GPU across multiple LLM models.
Compared with the uncompressed baseline, {\sys} achieves up to \textbf{16$\times$} attention speedup, \textbf{4.8$\times$} lower decode latency, and \textbf{7.3$\times$} higher throughput, while reducing GPU memory usage by \textbf{3$\times$}.
Despite these performance improvements, {\sys} preserves model accuracy, incurring only an average of \textbf{1.76\%} accuracy loss on the LongBench~\cite{bai_longbench_2024} and RULER~\cite{hsieh_ruler_2024} benchmarks.
This paper makes the following contributions:

\begin{itemize}
  \item A dynamic two-D compression method that applies both channel and sequence compression to the KV cache throughout the entire decode stage.
  It employs per-vector element selection and a dynamic compression strategy to preserve accuracy under high compression rates.

  \item A compressed KV cache management system that exploits underutilized GPU and CPU resources to manage the compressed KV cache and accelerate attention computation.

  \item A prototype implementation and evaluation.
  {\sys} achieves up to \textbf{16$\times$} attention speedup, \textbf{4.8$\times$} lower decode latency, and \textbf{7.3$\times$} higher throughput, while incurring only an average of \textbf{1.76\%} accuracy loss compared to the uncompressed baseline.

\end{itemize}

\section{Background and Motivation}
\label{sec:background}

\subsection{Large KV Cache in Long-Context LLM Inference}
\label{sec:bg:kvcache}

Large language models (LLMs) are increasingly deployed in scenarios that require extremely long contexts, including repository-level code generation~\cite{zhang_repocoder_2023,zhang_codeagent_2024}, multi-document question answering~\cite{wang_knowledge_2024,cheng_dualrag_2025}, and multi-turn agentic workflows~\cite{li_beyond_2026,wei_webagent-r1_2025}.
Frontier models~\cite{noauthor_llms_2026,noauthor_introducing_gpt5.4_nodate,noauthor_gemini3_2025,anthropic_whats_new_claude_46_2026,noauthor_LLaMA4_2025} now support 1M token context windows, e.g., Claude 4.6~\cite{anthropic_whats_new_claude_46_2026} and GPT-5.4~\cite{noauthor_introducing_gpt5.4_nodate}.
LLM inference consists of a {prefill} stage that processes input tokens and a {decode} stage that generates output tokens autoregressively.
To avoid redundant recomputation, modern inference frameworks maintain a KV cache to store key (K) and value (V) vectors of all historical tokens at each attention layer.

The KV cache resides in GPU memory to minimize computation latency.
However, its size scales linearly with the context length, making it the primary memory bottleneck.
For LLaMA-3.1-8B serving 8 concurrent requests at 128K tokens, the KV cache requires 128\,GB (8$\times$ the model weights); at 1M tokens, this escalates to 1\,TB (64$\times$ the model weights).
This memory pressure also increases attention latency: attention computation accounts for \textbf{84.7\%} and \textbf{97.9\%} of the total Time-Between-Tokens (TBT) at 128K and 1M contexts, respectively.

\subsection{Existing KV Cache Compression Methods}
\label{sec:bg:single}

Existing works~\cite{zhang_h2o_2023,tang_quest_2024,xu_think_2024,lee_infinigen_2024,sun_shadowkv_2025} compress the KV cache to reduce both memory footprint and latency along two dimensions: sequence compression and channel compression.

\textbf{Sequence Compression.}
Also referred to as sparse attention, this approach selects a subset of important tokens for attention computation.
It exploits the observation that the \emph{softmax} operation exponentially amplifies score differences, concentrating probability mass on a few high-scoring tokens.
As illustrated in Figure~\ref{fig:related}, existing methods compute attention scores via dot products between the current query and cached K vectors, retaining only the top-scoring tokens.
Quest~\cite{tang_quest_2024} accelerates this by computing scores at block granularity, while InfiniGen~\cite{lee_infinigen_2024} applies SVD to compress K vectors for fast selection.


\begin{figure}[t]
  \centering
  \includegraphics[width=\linewidth]{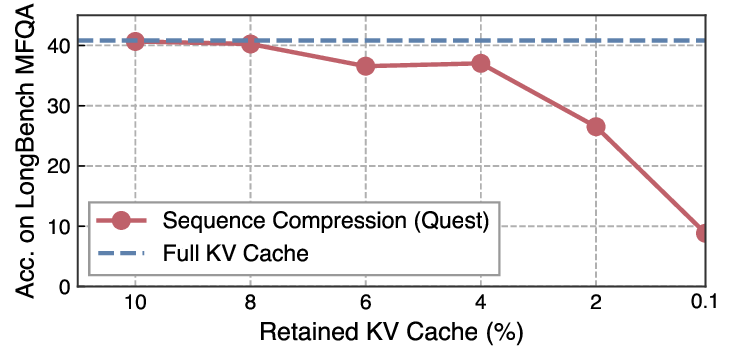}
    \caption{Model accuracy of sequence compression.}
  \label{fig:quest_acc}
  \vspace{-0.3cm}
\end{figure}

\textbf{Channel Compression.}
This approach compresses KV vectors along the channel dimension, exploiting the observation that outlier elements with large magnitudes are concentrated in a small subset of channels~\cite{hooper_kvquant_2024,xu_think_2024}.
These methods identify important channels across the KV cache and retain only these channels during attention computation, as shown in Figure~\ref{fig:related}.
For example, ThinK~\cite{xu_think_2024} computes a uniform channel mask from the prefill-stage K cache and applies it to compress all key vectors.
Other works~\cite{chang_palu_2024,sun_shadowkv_2025} leverage Singular Value Decomposition (SVD) to concentrate high-variance information into a small subset of channels, further improving compression quality.


\textbf{The Need for Two-Dimensional Compression.}
Context lengths have scaled exponentially from 64--128K to 1--10M tokens~\cite{noauthor_introducing_gpt5.4_nodate, noauthor_LLaMA4_2025}, yet single-dimensional compression cannot scale accordingly.
As shown in Figure~\ref{fig:quest_acc}, model accuracy degrades significantly as the sequence compression rate increases for Quest.
Moreover, dynamic sequence compression must retain the full KV cache in memory since any token may be selected at each decoding step, constraining the maximum batch size.
While offloading the KV cache to CPU or remote memory is feasible, it introduces substantial cross-device data transfer overhead and latency penalties~\cite{sheng_flexgen_2023,xiao_infllm_2024}.
Channel compression similarly cannot achieve higher compression rates without accuracy loss.
Combining both dimensions could enable higher compression rates for extended contexts.

Existing works have partially explored this direction.
InfiniGen~\cite{lee_infinigen_2024} uses channel compression to accelerate token selection but retains the full KV cache and uses uncompressed vectors for attention.
ThinK~\cite{xu_think_2024} applies channel compression only to K vectors, reporting significant accuracy loss when both dimensions are applied to the entire KV cache.
Consequently, no existing system achieves two-dimensional KV cache compression throughout the entire decode stage, encompassing both token selection and attention computation.


\subsection{Singular Value Decomposition (SVD)}
\label{sec:bg:svd}

Channel compression leverages Singular Value Decomposition (SVD) to concentrate significant information into a small number of top channels.
Given a matrix $K$, SVD decomposes it as $K = K' \cdot R$, where $K'$ is a rotated matrix with high-magnitude elements concentrated in the first few columns, and $R$ is computed via eigendecomposition of $K^T \cdot K$.
In practice, channel compression computes rotation matrices ($R_k$ and $R_v$) for the KV cache, applies the rotations ($K \cdot R_k^T$ and $V \cdot R_v^T$), and retains only the top-$r$ channels.
During attention, the rotated KV cache must be transformed back to the original space.

\section{Overview of {\sys}}
\label{sec:overview}

\subsection{Goals}
\label{sec:overview:goals}

The primary goal of {\sys} is to build a high-performance framework for serving long-context LLM inference.
By combining channel and sequence KV cache compression, {\sys} aims to achieve:
(1)~\textbf{Low Memory Usage}: both K and V caches are compressed along two dimensions. The memory reduction is sustained throughout token selection and attention computation;
(2)~\textbf{High Performance}: both decode latency and throughput are improved;
(3)~\textbf{High Accuracy}: accuracy loss is minimized.

\begin{figure}[tbp]
  \centering
  \includegraphics[width=\columnwidth]{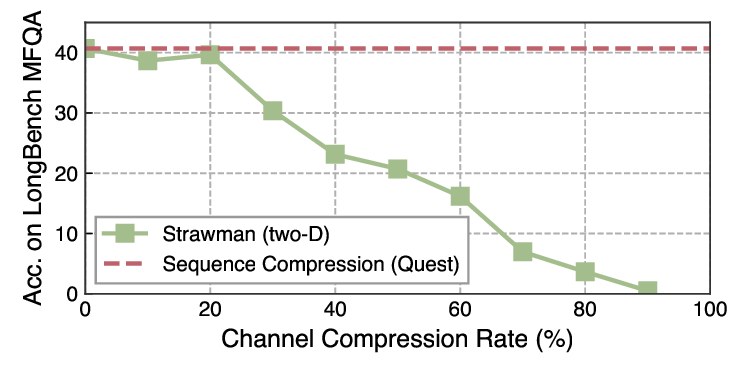}
  \caption{Accuracy of naively applying channel compression to sequence compression.}
  \label{fig:quest_svd_acc}
  \vspace{-0.1cm}
\end{figure}

\subsection{Challenges and Approaches}
\label{sec:overview:challenges}

To further reduce the size of the KV cache, a straightforward approach is to combine channel and sequence compression.
However, realizing the full benefit of such two-D (dimensional) compression is non-trivial, presenting the following challenges:

\begin{figure}[tbp]
  \centering
  \includegraphics[width=\columnwidth]{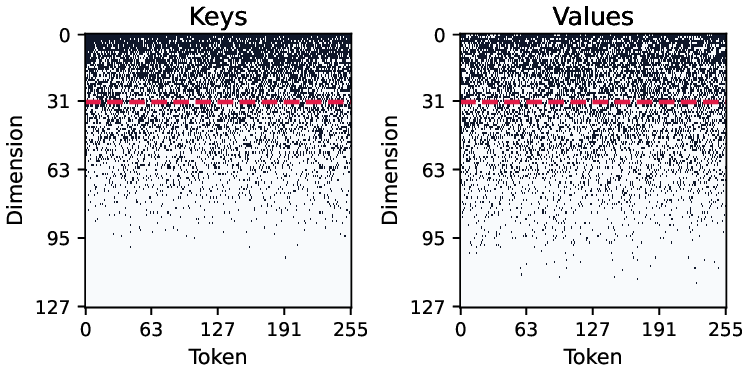}
  \caption{Distribution of top-25\% elements in KV cache applied with SVD. (Red line: top-25\% channels)}
  \label{fig:kv_cache_visualization}
  \vspace{-0.1cm}
\end{figure}

\textbf{Challenge 1: Accuracy Challenge.}
\label{sec:overview:accuracy}
Naively applying channel compression to sequence compression introduces significant accuracy loss.
We propose a strawman design that applies channel compression to Quest, a dynamic sequence compression method, as shown in Figure~\ref{fig:related}.
It applies SVD rotation to the KV cache, selects the top-$r$ important channels for both K and V vectors, and then selects the top 10\% important token blocks for attention computation.
We evaluate this strawman on LongBench~\cite{bai_longbench_2024} by varying the channel compression rate (Figure~\ref{fig:quest_svd_acc}).
Compared to sequence compression alone (Quest), the strawman significantly degrades accuracy, with accuracy loss reaching 24.5\% at a 30\% channel compression rate and 82.8\% at 70\%.

\textbf{\emph{Observation 1: Non-Uniform Importance Distribution.}}
We observe that the importance distribution within the KV cache is non-uniform.
However, existing compression methods often apply a one-size-fits-all strategy to the entire KV cache.
For example, ThinK~\cite{xu_think_2024} applies a uniform channel mask to all tokens, and InfiniGen~\cite{lee_infinigen_2024} applies a uniform SVD rotation matrix to all requests, as shown in Figure~\ref{fig:related}.
Such uniform strategies discard a significant number of important elements, thereby degrading model accuracy.
Specifically, the importance distribution exhibits two levels of non-uniformity.
First, the distribution of important elements varies significantly across individual KV vectors.
As shown in Figure~\ref{fig:kv_cache_visualization}, only 62.28\% of the top-25\% important elements (marked in black) reside in the globally top-25\% important channels (above the red line) for a small context of 256 tokens, even after SVD is applied.
Second, the element distribution characteristics vary across different KV cache regions.
For example, in some regions, element magnitudes exhibit high variance with pronounced outliers, while in others, elements are distributed more uniformly with similar magnitudes.
The former is amenable to a high compression rate, whereas the latter requires a lower rate.
Moreover, applying a single SVD rotation matrix to the entire KV cache cannot efficiently cluster important elements, compared to using a separate rotation matrix per region.

\textbf{\emph{Approach-1: Dynamic Two-D Compression.}}
Based on the above observations, {\sys} introduces dynamic two-D compression, which performs two-dimensional compression with adaptive strategies, to address the accuracy challenge (Section~\ref{sec:trim}).
Unlike existing works that apply a uniform channel mask to all KV vectors, {\sys} first introduces \emph{per-vector element selection}, which identifies important elements for each KV vector individually.
Furthermore, {\sys} introduces a \emph{dynamic compression strategy} that partitions the KV cache into multiple segments and applies the most suitable strategy to each segment.
The dynamic two-D compression enables {\sys} to apply two-dimensional KV cache compression throughout the entire decode stage, increasing the compression rate while preserving model accuracy.




\textbf{Challenge 2: Performance Challenge.}
\label{sec:overview:efficiency}
Although {\sys}'s dynamic two-D compression achieves a high KV cache compression rate, it introduces fine-grained sparse memory access patterns and substantial memory management overhead, which can degrade performance to below the uncompressed baseline.
First, as shown in Figure~\ref{fig:method}, {\sys} generates a fine-grained and unstructured sparse KV cache matrix, whereas modern GPUs are optimized for dense tensor operations.
Such a sparse matrix exhibits poor cache locality and significantly increases the GPU cache miss rate.
Theoretically, although a 75\%-compressed matrix reduces memory accesses by 75\%, it may increase the cache miss rate by $4\times$, negating the speedup from compression.
Additionally, handling the sparsity (e.g., skipping zero values) introduces further overhead, making the attention latency with such a sparse KV cache substantially higher than with the uncompressed full KV cache.
We measure the attention latency using a 25\%--80\%-compressed KV cache generated by {\sys}'s method with cuSPARSE~\cite{cusparse}, and find it 25--136$\times$ higher than the uncompressed baseline.
This overhead also stems from the format conversion required by cuSPARSE, which is designed for extreme sparsity, e.g., 99\% of values are zero.
Second, heavy KV cache management operations are introduced into the critical decode stage.
On one hand, KV vectors need to be compressed and decompressed during inference.
On the other hand, the dynamic compression strategy introduces memory partitioning, SVD computation, and recompression operations for newly generated KV cache segments.
These operations further add overhead to the decode stage.

\textbf{\emph{Observation 2: Imbalanced Resource Utilization.}}
For long-context requests, attention computation dominates both memory usage and latency.
We observe that decode-stage attention exhibits a highly imbalanced workload: GPU memory bandwidth is fully saturated, while GPU compute units (especially CUDA cores) and the CPU remain significantly underutilized, even with highly optimized operators~\cite{dao2022flashattention, dao2023flashattention, NEURIPS2024_7ede97c3, ye2025flashinfer}.
Profiling FlashInfer~\cite{ye2025flashinfer} during the decode stage of a 256K-token request on LLaMA-3.1-8B with an A800 GPU, we find that memory bandwidth utilization reaches 90.5\%, while CUDA core utilization is only 10.35\% and almost all CPU cores are idle.

\textbf{\emph{Approach-2: Compressed KV Cache Management.}}
Based on the above observation, {\sys} introduces a novel compressed KV cache management method that exploits underutilized resources to manage the compressed KV cache and accelerate attention computation, thereby addressing the performance challenge (Section~\ref{sec:system}).
{\sys} first introduces \emph{packed sparse attention} to accelerate attention computation over the two-D compressed KV cache.
Furthermore, \emph{heterogeneous double compression buffering} is introduced to reduce the overhead of compressed KV cache management.

\subsection{System Overview}
\label{sec:overview:solution}

{\sys} is a novel framework that exploits the non-uniform importance distribution of the KV cache to perform two-dimensional compression along both the channel and sequence dimensions.
As introduced above, {\sys} employs a dynamic two-D compression method to achieve both a high KV cache compression rate and model accuracy (Section~\ref{sec:trim}).
It further introduces a compressed KV cache management method to realize the performance benefits of compression (Section~\ref{sec:system}).

\textbf{Dynamic Two-D Compression.}
{\sys} combines per-vector element selection with dynamic token selection to perform two-dimensional compression.
Specifically, it introduces a \textbf{two-stage two-D compression} method (Section~\ref{sec:trim:two-stage}).
Stage~1 performs per-vector channel compression during both the prefill and decode stages for input and output tokens.
It applies SVD rotation to the KV cache and selects the top-$r$ important elements for each token's KV vectors.
Stage~2 performs dynamic sequence compression during the decode stage.
Before each attention layer, it selects the top-$k$ tokens from the channel-compressed KV cache using the current query vector.
Additionally, the \textbf{dynamic compression strategy} partitions the KV cache into multiple segments and applies the most suitable strategy to each segment (Section~\ref{sec:trim:generated}).

\textbf{Compressed KV Cache Management.}
{\sys} leverages underutilized GPU CUDA cores and the CPU to accelerate attention computation and manage the compressed KV cache.
First, {\sys} introduces packed sparse attention to accelerate attention computation over the two-D compressed KV cache by leveraging underutilized CUDA cores.
It introduces a \textbf{packed KV format} to represent the sparse KV cache in a dense format, and a \textbf{compressed KV encoding} to efficiently perform stage-1 compression and convert the results to the packed KV format during the prefill stage.
This packed format improves cache locality for the sparse KV cache and reduces the GPU cache miss rate.
During the decode stage, \textbf{PackedAttention} is introduced to leverage underutilized CUDA cores for performing stage-2 compression and attention computation with the packed KV cache (Section~\ref{sec:system:kernels}).

Furthermore, {\sys} introduces \textbf{heterogeneous double compression buffering} to reduce the overhead of dynamic KV cache compression and management by leveraging the underutilized CPU (Section~\ref{sec:system:segmented}).
Once a new token is generated, it is temporarily compressed and stored in the GPU-side compression buffer for subsequent decode computation in the interim.
Meanwhile, a CPU-side compression buffer is used to perform KV cache management operations asynchronously, including segment partitioning and compression strategy generation.
Once a new segment is fully generated and compressed, a switching mechanism replaces the temporarily compressed KV segment with the final compressed version without blocking the decode stage.
An incremental method is proposed to accelerate compression strategy generation (Section~\ref{sec:system:segmented}).


\section{Dynamic Two-D Compression}
\label{sec:trim}

\subsection{Basic Method}
\label{sec:trim:overview}

\begin{figure}[t]
  \centering
  \includegraphics[width=\linewidth]{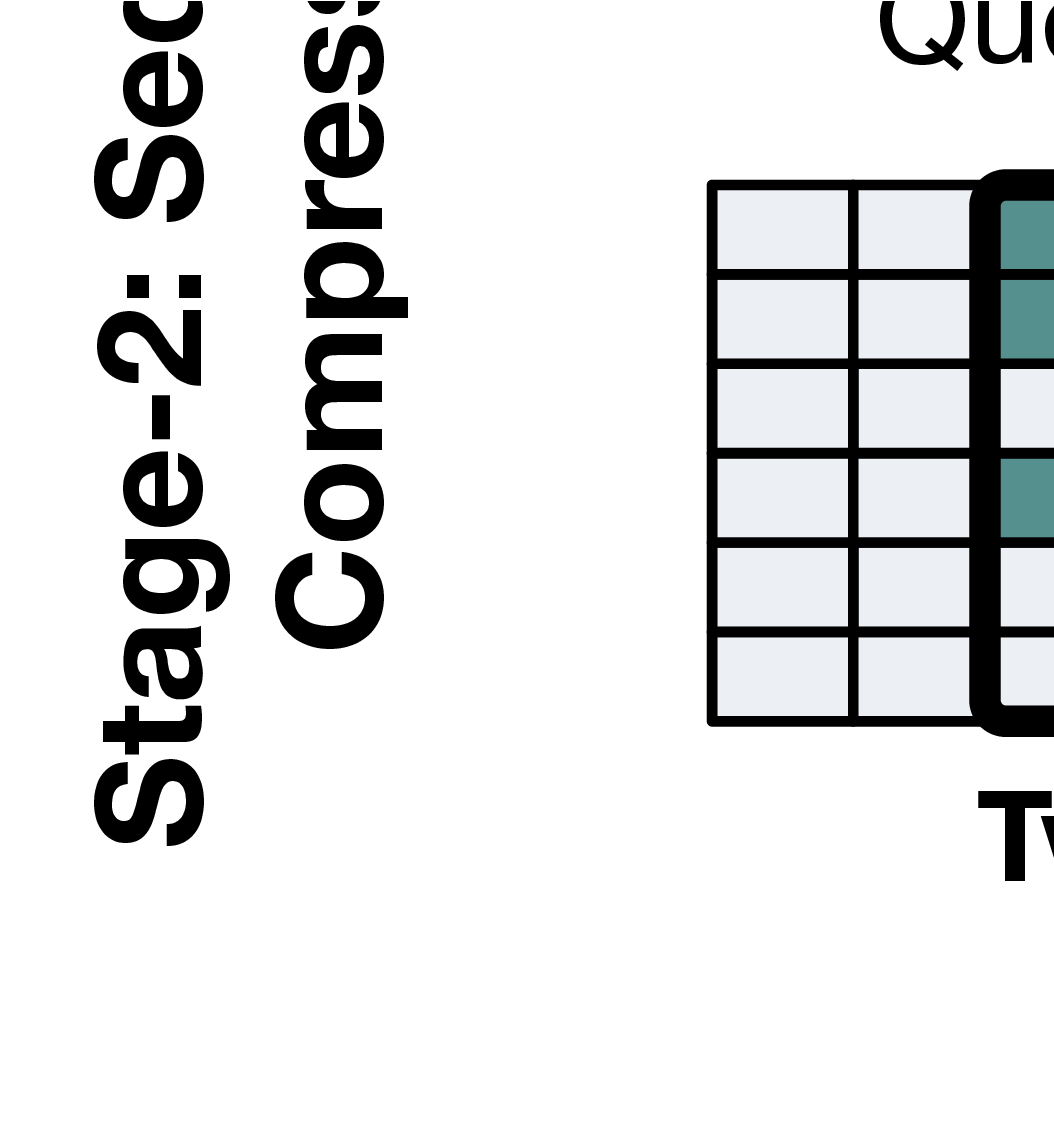}
  \caption{Dynamic Two-D Compression.}
  \label{fig:method}
\end{figure}

{\sys} introduces dynamic two-D compression to exploit the non-uniform importance distribution of the KV cache with dynamic compression strategies.
Figure~\ref{fig:method} presents an overview.
The entire KV cache is partitioned into multiple segments, and {\sys} performs a two-stage two-D compression on both the K and V vectors for each segment (Section~\ref{sec:trim:two-stage}).
The compressed KV caches from all segments are then passed to the attention computation.
At runtime, {\sys} dynamically partitions the KV cache and assigns a suitable compression strategy to each segment (Section~\ref{sec:trim:generated}), aiming for a better balance between performance and accuracy.

\begin{figure*}[t]
  \centering
  \includegraphics[width=0.85\linewidth]{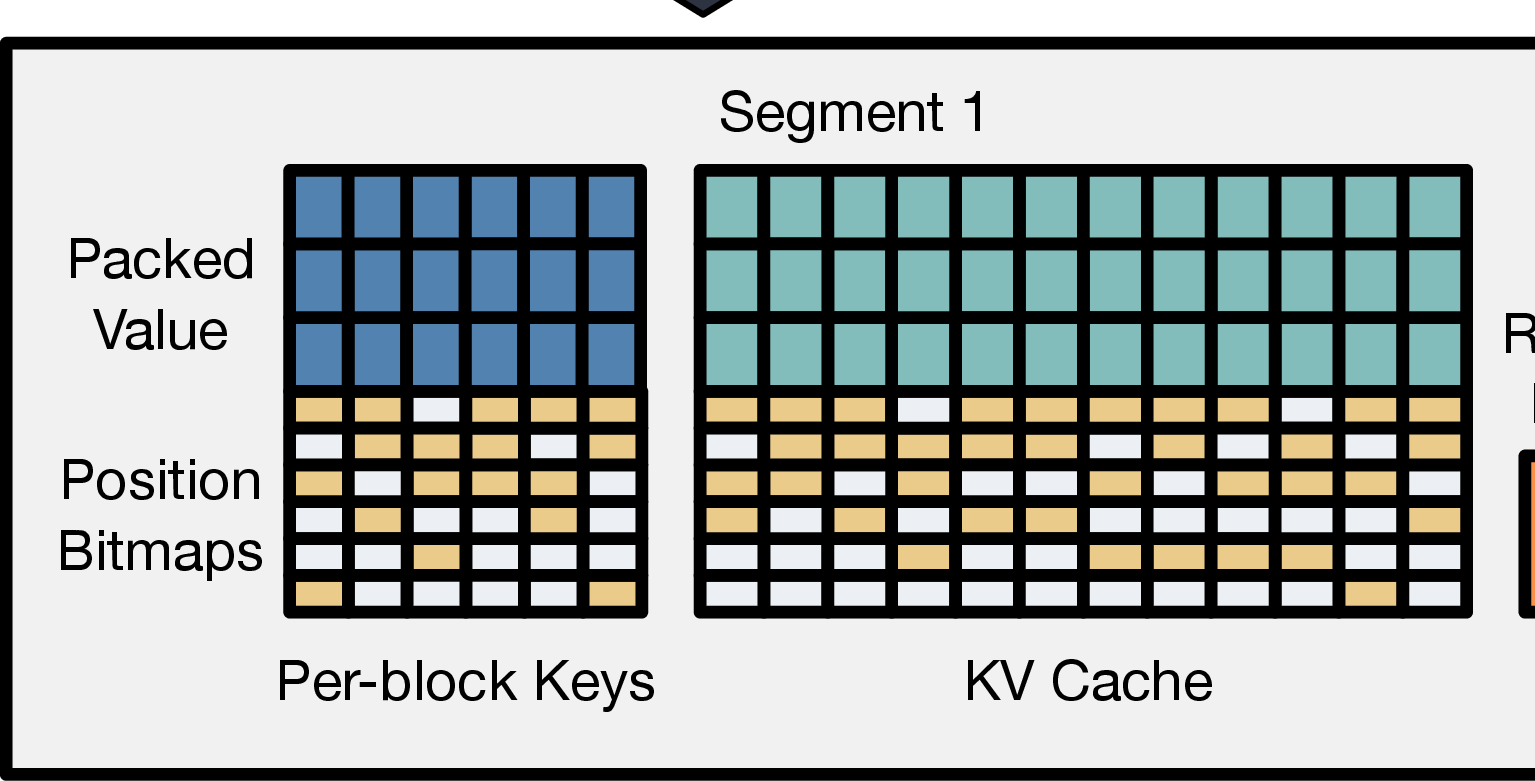}
  \caption{\textbf{Overview of Packed Sparse Attention:} including packed KV format, flexible KV encoding and PackedAttention.}
  \label{fig:overview}
\end{figure*}

\subsection{Two-Stage Two-D Compression}
\label{sec:trim:two-stage}

The two-stage compression consists of per-vector channel compression (stage~1) and dynamic sequence compression (stage~2).
Stage~1 is performed during both the prefill and decode stages for input and output tokens, respectively, while stage~2 is performed during the decode stage before each attention layer.

\textbf{Stage-1: Per-Vector Channel Compression.}
Stage-1 compression selects the top-$r$ important elements for each KV vector.
It first computes the SVD rotation matrix $R$ for the current segment's K and V matrices, respectively, and rotates both matrices by their corresponding $R$ to cluster important elements.
For each KV vector, the top-$r$ elements with the largest absolute values are selected and all remaining elements are discarded.
For input tokens, whose KV segments are available after the prefill stage, compression is applied directly (Section~\ref{sec:system:lean}).
For output tokens, which are generated one by one, {\sys} introduces heterogeneous double compression buffering to perform channel compression incrementally (Section~\ref{sec:system:segmented}).

\textbf{Stage-2: Dynamic Sequence Compression.}
Stage-2 compression dynamically selects the top-$k$ important tokens before each attention layer during the decode stage.
For each KV cache segment, {\sys} partitions all tokens into small blocks (e.g., 8 or 16 tokens) and computes a per-block key vector as the mean of the key vectors within each block.
The per-block keys are also channel-compressed together with the original K vectors during stage-1.
During stage-2, an attention score is computed for each block via the dot product between the current query vector and the per-block key, and the top-$k$ blocks with the highest scores are selected for attention computation.
Benefiting from stage-1, both the per-block key vectors and the selected tokens' KV vectors are channel-compressed.

\subsection{Dynamic Compression Strategy}
\label{sec:trim:generated}

{\sys} partitions the entire KV cache into multiple segments and assigns a distinct compression strategy to each.
Input and output tokens are partitioned separately: input tokens' KV cache is split into configurable fixed-size segments (e.g., 64K tokens) during the prefill stage to minimize partitioning overhead, while output tokens' KV cache is dynamically partitioned during the decode stage.
For output tokens, {\sys} maintains a \emph{compression loss} quantifying the quality degradation of the current compression strategy on newly generated KV vectors.
As each new token is generated, it is appended to the current segment and the compression loss is updated; once the loss exceeds a predefined threshold, {\sys} finalizes the current segment and initiates a new one.
A maximum segment length (64K tokens) is imposed to ensure periodic refresh even when the compression loss remains stable.

\textbf{Compression Strategy.}
Each segment's compression strategy includes parameters for both channel and sequence compression.
The channel compression parameters include:
1) SVD rotation matrix, used to rotate the segment's K and V matrices, respectively;
2) channel compression rate, determining how many elements are discarded per vector;
3) element group size, specifying the granularity at which neighboring elements are selected together (Section~\ref{sec:system:bitmap}).
The sequence compression parameters include:
1) sequence compression rate, determining how many tokens are discarded per block;
2) token block size, specifying the granularity at which neighboring tokens are selected together.
Section~\ref{sec:system:segmented} describes how {\sys} partitions the KV cache into segments and generates the compression strategy.

\section{Compressed KV Cache Management}
\label{sec:system}

\subsection{Packed KV Format}
\label{sec:system:bitmap}

Both the KV vectors and per-block keys are channel-compressed and stored in GPU memory as fine-grained, unstructured sparse matrices that cannot be efficiently processed by GPUs.
{\sys} introduces the packed KV format to convert them into a dense representation, which is subsequently used by PackedAttention (Section~\ref{sec:system:kernels}).
Let $D$ denote the number of channels of the original vector and $r$ denote the number of retained elements.
As shown in Figure~\ref{fig:method}, {\sys} packs all retained elements contiguously for each sparse vector and employs a $D$-bit position bitmap, where each bit indicates whether the corresponding element is retained, rather than storing per-element indices as in conventional COO/CSR formats.



\textbf{Variable-Granularity Bitmap.}
After compacting the sparse KV cache, the position bitmap occupies substantial memory.
For a KV vector with a 75\% compression rate and 8-bit elements, the packed elements occupy 25\% of the original memory, while the bitmaps account for 12.5\%.
To reduce this overhead, {\sys} selects groups of adjacent elements together rather than individual elements, using one bit per element group.
The dynamic compression strategy (Section~\ref{sec:trim:generated}) selects a suitable element group size (1, 2, or 4) for each KV cache segment.

\subsection{Compressed KV Encoding}
\label{sec:system:lean}

During the prefill stage, the compressed KV encoding performs channel compression and converts the compressed KV cache into the packed KV format.
Section~\ref{sec:system:segmented} describes how the output tokens' KV cache is channel-compressed and encoded.
Each input KV cache segment has a distinct compression strategy and is encoded separately.
Once the KV cache is computed, {\sys} calculates the SVD rotation matrices for the K and V matrices and rotates them accordingly.
Meanwhile, all tokens are divided into multiple token blocks, with the block size specified by the strategy.
A per-block key is computed as the average of the key vectors within each block.
{\sys} then performs channel compression on the K, V, and per-block key matrices, and stores the results in the packed KV format in GPU memory (Figure~\ref{fig:overview}).

\textbf{Channel Truncation.}
{\sys} observes that globally unimportant channels consistently exhibit low values across tokens after SVD rotation.
As shown in Figure~\ref{fig:kv_cache_visualization}, the bottom 25\% channels (by importance) are never selected.
Accordingly, {\sys} truncates these channels, which need not be represented in the position bitmap, accelerating channel compression and reducing bitmap memory overhead.

\textbf{Asynchronous Encoding.}
During the prefill stage, the input tokens' KV cache is computed layer by layer.
The compressed KV encoding is overlapped with the prefill computation.
Encoding $layer_{i-1}$'s KV cache runs concurrently with the normal computation of $layer_i$.
The encoding overhead is evaluated in Section~\ref{sec:eval}.

\begin{figure}[t]
  \centering
  \includegraphics[width=\linewidth]{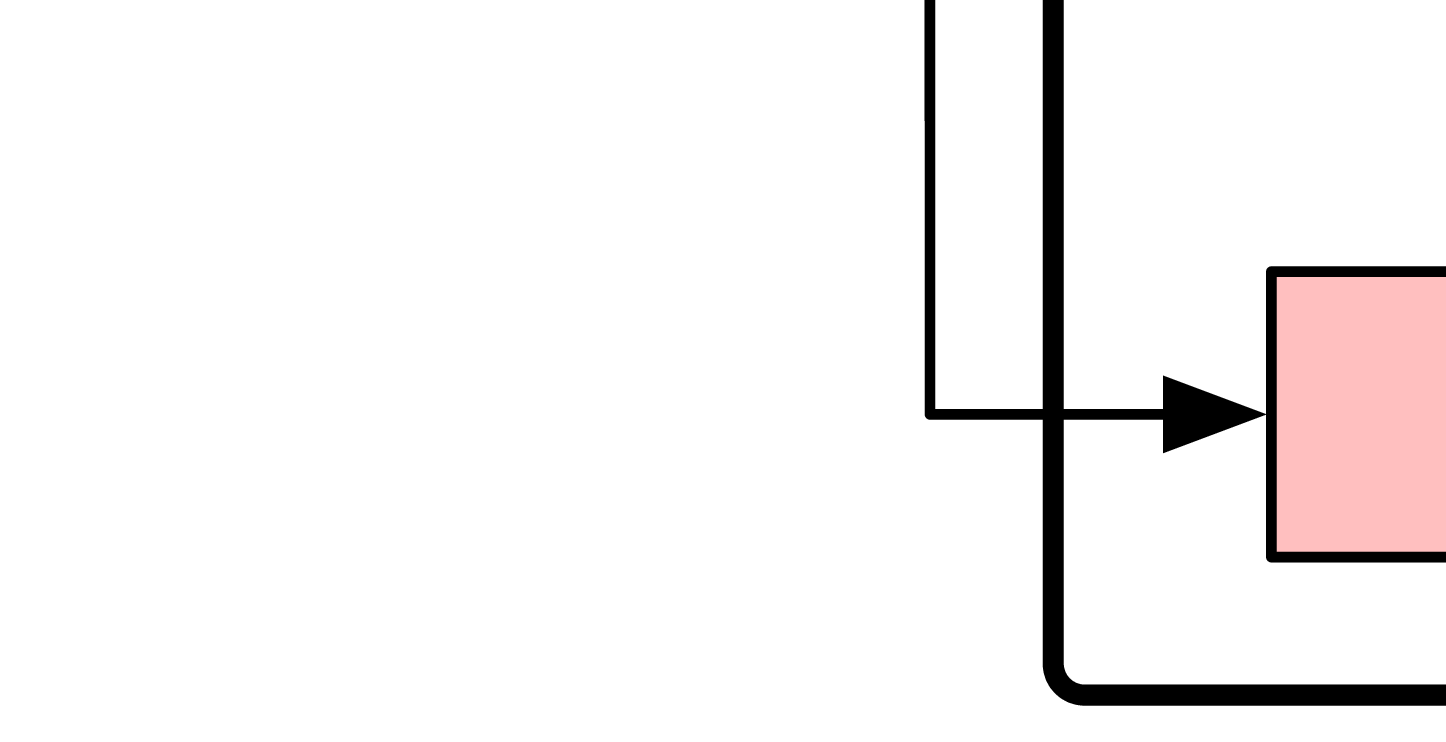}
  \caption{Heterogeneous Double Compression Buffering.}
  \label{fig:segmented}
\end{figure}

\subsection{PackedAttention}
\label{sec:system:kernels}

PackedAttention performs sequence compression and attention computation in each attention layer of the decode stage.
As shown in Figure~\ref{fig:overview}, it computes the attention score for each token block, selects the top-$k$ blocks, and performs attention computation with the two-D-compressed KV cache.
The attention score computation and the attention computation are the two main bottlenecks.
Both computations must access the channel-compressed and packed per-block keys and KV caches.
Naively, these vectors would need to be unpacked and decompressed before use, significantly degrading performance.
PackedAttention eliminates these overheads.


First, PackedAttention operates directly on packed vectors without unpacking them.
In both the attention score and attention computation, the packed K, V, and per-block key matrices must be dot-producted with another vector, e.g., the query vector.
Instead of unpacking and using Tensor cores, {\sys} leverages CUDA cores to operate directly on packed vectors.
As shown in the top-right of Figure~\ref{fig:overview}, a packed vector product computes the dot product between a dense vector (e.g., the query) and a packed sparse vector (e.g., a key) by reconstructing element positions on-the-fly.
For each set bit in the position bitmap, {\sys} retrieves the element index, fetches the corresponding element from the query vector, and performs the multiplication; all products are accumulated to produce the final result.
To accelerate this computation, {\sys} partitions the position bitmap into multiple chunks and assigns each chunk to a thread.
This approach exhibits good cache locality and reduces the GPU cache miss rate when accessing sparse vectors, yielding significant performance improvements from per-vector channel compression.

Second, PackedAttention merges the decompression operation into the single uncompressed vector rather than applying it to the entire compressed matrix.
Taking the attention score computation as an example, the query vector must be dot-producted with the channel-compressed per-block key matrix, which is SVD-rotated.
It must be rotated back before the dot product.
Instead of rotating the large matrix, {\sys} merges the rotation into the single query vector and uses the rotated query to compute the attention scores.
This approach is mathematically equivalent and offers substantially better performance.

\textbf{Compatibility with FlashAttention.}
PackedAttention is implemented on top of FlashAttention~\cite{dao2022flashattention}, which processes the KV cache in small chunks (e.g., 8 tokens), allowing unselected chunks to be directly skipped.
The standard dot product is replaced with the packed vector product.
Although PackedAttention applies different compression strategies to different KV cache segments, the segment size is substantially larger than the chunk size, so this incurs no additional overhead.

\textbf{KV Cache Update Buffer.}
For the KV vectors of newly generated tokens, {\sys} uses an update buffer to temporarily hold them without any compression, as shown in Figure~\ref{fig:overview}.
Once the buffer is full, all KV vectors within it are channel-compressed and packed, and per-block keys are computed, following the same procedure as the compressed KV encoding (Section~\ref{sec:system:lean}).
In our current implementation, the buffer size is 32 tokens, which exceeds the maximum block size.


\begin{table*}[t]
  \centering
  \caption{Model accuracy with RULER and LongBench.}
  \label{tab:accuracy}
  \small
  \setlength{\tabcolsep}{4pt}
  \setlength{\aboverulesep}{2pt}
  \setlength{\belowrulesep}{2pt}
  \renewcommand{\arraystretch}{0.85}
  \resizebox{\textwidth}{!}{%
  \begin{tabular}{l|l|rrrrrrrrrrrr|r}
    \toprule
    & & \multicolumn{13}{c}{\textbf{RULER}} \\
    \cmidrule(lr){3-15}
    \textbf{Model} & \textbf{Method}
      & \textbf{S1} & \textbf{S2} & \textbf{S3} & \textbf{MK1} & \textbf{MK2} & \textbf{MK3}
      & \textbf{MQ} & \textbf{MV} & \textbf{QA-1} & \textbf{QA-2} & \textbf{VT} & \textbf{FWE}
      & \textbf{Avg} \\
    \midrule
    \multirow{3}{*}{\makecell[l]{LLaMA-3.1-8B}}
      & Full  & 100.00 & 100.00 & 100.00 & 100.00 & 98.00 & 96.00 & 100.00 & 98.00 & 56.00 & 62.83 & 99.20 & 88.67 & 91.56 \\
      & Quest & 100.00 & 100.00 & 100.00 & 98.00 & 98.00 & 88.00 & 100.00 & 98.50 & 54.00 & 61.50 & 96.40 & 90.67 & 90.42 \\
      & \sys  & 100.00 & 100.00 & 100.00 & 98.00 & 98.00 & 92.00 & 97.50 & 99.50 & 52.00 & 58.83 & 96.40 & 86.00 & 89.85 \\
    \midrule
    \multirow{3}{*}{\makecell[l]{Qwen3-4B}}
      & Full  & 100.00 & 100.00 & 100.00 & 100.00 & 100.00 & 100.00 & 100.00 & 100.00 & 70.00 & 54.17 & 99.20 & 92.00 & 92.95 \\
      & Quest & 100.00 & 100.00 & 100.00 & 100.00 & 100.00 & 98.00 & 98.50 & 99.00 & 68.00 & 54.83 & 99.20 & 88.00 & 92.13 \\
      & \sys  & 100.00 & 100.00 & 98.00 & 100.00 & 98.00 & 100.00 & 99.00 & 98.00 & 58.00 & 54.17 & 91.20 & 89.30 & 90.47 \\
    \midrule
    \multirow{3}{*}{\makecell[l]{Ministral-3-8B}}
      & Full  & 100.00 & 100.00 & 100.00 & 98.00 & 98.00 & 98.00 & 96.00 & 98.50 & 56.00 & 52.11 & 100.00 & 84.00 & 90.05 \\
      & Quest & 100.00 & 100.00 & 100.00 & 98.00 & 98.00 & 98.00 & 96.50 & 100.00 & 54.00 & 50.83 & 100.00 & 76.67 & 89.33 \\
      & \sys  & 100.00 & 100.00 & 100.00 & 90.00 & 98.00 & 94.00 & 90.00 & 96.50 & 50.00 & 48.11 & 100.00 & 76.67 & 86.94 \\
    \midrule\midrule
    & & \multicolumn{13}{c}{\textbf{LongBench}} \\
    \cmidrule(lr){3-15}
    \textbf{Model} & \textbf{Method}
      & \textbf{LCC} & \textbf{SAM}  & \textbf{TQA}
      & \textbf{2Wiki} & \textbf{Hotp} & \textbf{MUS}
      & \textbf{MFQA} & \textbf{NQA} & \textbf{QAS}
      & \textbf{MN} & \textbf{QM}
      & \textbf{PC}
      & \textbf{Avg} \\
    \midrule
    \multirow{3}{*}{\makecell[l]{LLaMA-3.1-8B}}
      & Full  & 33.15 & 44.26  & 92.29 & 24.23 & 32.51 & 15.90 & 40.82 & 31.37 & 28.02 & 26.77 & 24.22 & 8.57  & 33.51 \\
      & Quest & 32.37 & 44.25  & 92.28 & 23.37 & 34.63 & 16.44 & 40.65 & 30.66 & 20.94 & 25.21 & 23.63 & 7.06  & 32.62 \\
      & \sys  & 32.87 & 43.71  & 90.16 & 24.62 & 37.78 & 18.96 & 35.60 & 30.87 & 24.34 & 25.56 & 23.66 & 6.00  & 32.84 \\
    \midrule
    \multirow{3}{*}{\makecell[l]{Qwen3-4B}}
      & Full  & 5.25 & 46.96  & 88.01 & 11.38 & 12.44 & 7.14 & 28.19 & 9.08 & 18.87 & 21.91 & 18.50 & 3.60  & 22.61 \\
      & Quest & 5.07 & 45.75  & 88.44 & 12.08 & 12.17 & 6.74 & 29.32 & 11.47 & 20.51 & 21.44 & 18.68 & 3.56  & 22.94 \\
      & \sys  & 5.10 & 45.48  & 88.58 & 11.66 & 11.65 & 6.69 & 28.46 & 10.89 & 14.23  & 21.29 & 18.67 & 3.37  & 22.17 \\
    \midrule
    \multirow{3}{*}{\makecell[l]{Ministral-3-8B}}
      & Full  & 25.90 & 25.89  & 89.87 & 11.82 & 10.61 & 8.32 & 35.02 & 13.53 & 15.23 & 12.87 & 23.71 & 4.50  & 23.11 \\
      & Quest & 26.43 & 25.54  & 89.87 & 12.77 & 10.83 & 8.23 & 35.26 & 13.76 & 16.03 & 12.18 & 24.12 & 4.00  & 23.25 \\
      & \sys  & 25.80 & 24.96  & 89.53 & 11.51 & 9.91 & 7.38 & 34.31 & 14.17 & 20.30 &17.55 & 23.26 & 2.50  & 23.43 \\
    \bottomrule
  \end{tabular}
  }
\end{table*}

\subsection{Heterogeneous Double Compression Buffering}
\label{sec:system:segmented}

{\sys} assigns a compression strategy to each KV cache segment, but the strategy can only be generated after the entire segment is determined.
However, the decode stage generates tokens one by one, and leaving a large non-terminated segment (e.g., 32K tokens) uncompressed is inefficient.
Heterogeneous double compression buffering addresses this by leveraging the underutilized CPU for segment partitioning, strategy generation, and compression.

As shown in Figure~\ref{fig:segmented}, a GPU-side compression buffer temporarily holds the KV cache of the current segment (segment$_i$ in the figure).
It is compressed using the previous segment's strategy since the current segment has not yet been finalized.
Meanwhile, all KV vectors of newly generated tokens are copied to the CPU-side compression buffer without compression.
Using this buffer, the CPU computes the compression loss of compressing segment$_{i}$ with the previous strategy.
Once the loss exceeds a predefined threshold, the CPU finalizes the current segment, computes a new compression strategy, and performs stage-1 compression following the same procedure as the compressed KV encoding (Section~\ref{sec:system:lean}).

All CPU-side operations run asynchronously without blocking the GPU-side decode stage.
Once segment$_{i}$ is compressed, {\sys} replaces the temporarily compressed GPU-side segment with the final CPU-side result, and the new strategy is transferred to the GPU.

\subsection{Incremental Strategy Generation}
\label{sec:system:incremental}

{\sys} executes most KV cache management operations, including segment partitioning, SVD computation, strategy generation, and compression, on the CPU.
An incremental mechanism is introduced for segment partitioning and SVD rotation matrix generation.
Rather than processing the entire segment at once, {\sys} performs these operations incrementally for each newly generated token.

For segment partitioning, {\sys} computes the compression loss for each token's KV vectors by comparing the channel-compressed vectors with the original ones, and maintains an accumulated total loss.
Once the total loss exceeds a predefined threshold, the current segment is finalized.
For the SVD rotation matrix, which is derived from the covariance matrix $\Sigma$ of a segment's KV matrix via eigendecomposition.
{\sys} incrementally updates $\Sigma$ using each new token's outer product.

With this mechanism, {\sys} can rapidly determine whether the current segment should be finalized, and the SVD rotation matrix can be directly computed from the up-to-date $\Sigma$.
Subsequently, {\sys} selects suitable strategy parameters for the newly terminated segment.
The SVD rotation matrices are already computed at this point.
All other parameters are selected from several candidate values.
For channel compression, {\sys} first reorders the rotated matrices so that channels with similar selection patterns become neighbors.
Then, {\sys} evaluates the compression loss for each candidate compression rate (e.g., 87.5\%, 75\%, and 62.5\%) and element group size (e.g., 1, 2, 4).
The highest compression rate and largest group size whose loss remains below a threshold are selected.
For sequence compression, {\sys} computes the intra-block key variance for multiple candidate block sizes (e.g., 4, 8, 16) and selects the largest size whose variance remains below a threshold.
This computation can also be performed incrementally for each newly generated token block.
The sequence compression rate is statically determined in our current implementation.

\section{Evaluation}
\label{sec:eval}

We implement a prototype of {\sys} based on PyTorch~\cite{pytorch2019} with approximately 5,300 lines of Python and 2,800 lines of CUDA.
The evaluation aims to answer the following questions:

\begin{itemize}
  \item \textbf{Q1}: How does {\sys} affect model accuracy on long-context tasks?
  \item \textbf{Q2}: How effectively does {\sys} reduce the KV cache memory footprint?
  \item \textbf{Q3}: What is the attention latency of {\sys}?
  \item \textbf{Q4}: What is the end-to-end serving performance of {\sys}?
\end{itemize}


\subsection{Setup}
\label{sec:eval:setup}

We evaluate the accuracy and performance of {\sys} on an NVIDIA H800 GPU equipped with 80GB HBM.
We compare {\sys} against two baselines:
1) Full baseline, which does not perform any KV cache compression.
For the microbenchmark, it employs a FlashAttention-based attention kernel.
For the end-to-end evaluation, it uses vLLM~\cite{kwon_efficient_2023}.
2) Quest~\cite{tang_quest_2024}, which performs sequence-only compression.
LLaMA-3.1-8B~\cite{noauthor_LLaMA_2024} is used in our performance evaluation, and accuracy is further evaluated with Qwen3-4B~\cite{yang_qwen3_2025} and Ministral-3-8B~\cite{liu_ministral_2026}.
By default, Quest uses a block size of 16 while {\sys} uses a block size of 8.
Both methods select 10\% of tokens.
{\sys} is configured to retain 25\% of channels.
The floating-point precision is BF16 for all systems.

\subsection{Accuracy}
\label{sec:eval:accuracy}

\begin{figure*}[ht]
  \centering
  \includegraphics[width=0.8\linewidth]{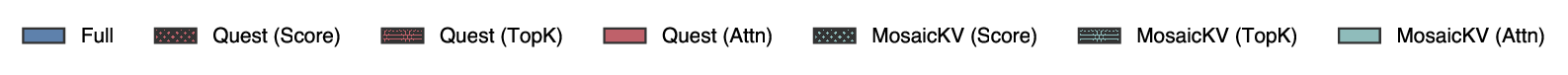}
  \vspace{1pt}

  \begin{minipage}[t]{0.33\linewidth}
    \centering
    \includegraphics[width=\linewidth]{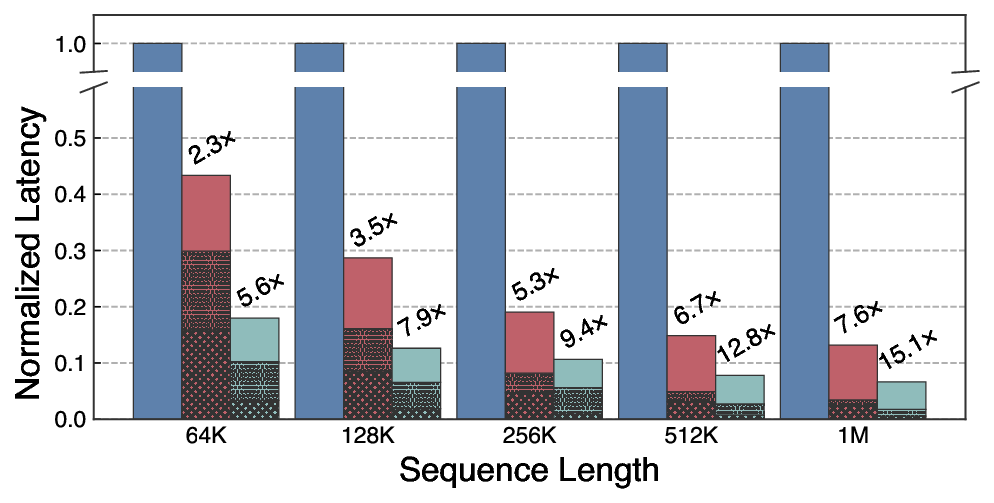}
    \vspace{-20pt}
    \caption*{(a) Batch size = 1}
  \end{minipage}\hfill
  \begin{minipage}[t]{0.33\linewidth}
    \centering
    \includegraphics[width=\linewidth]{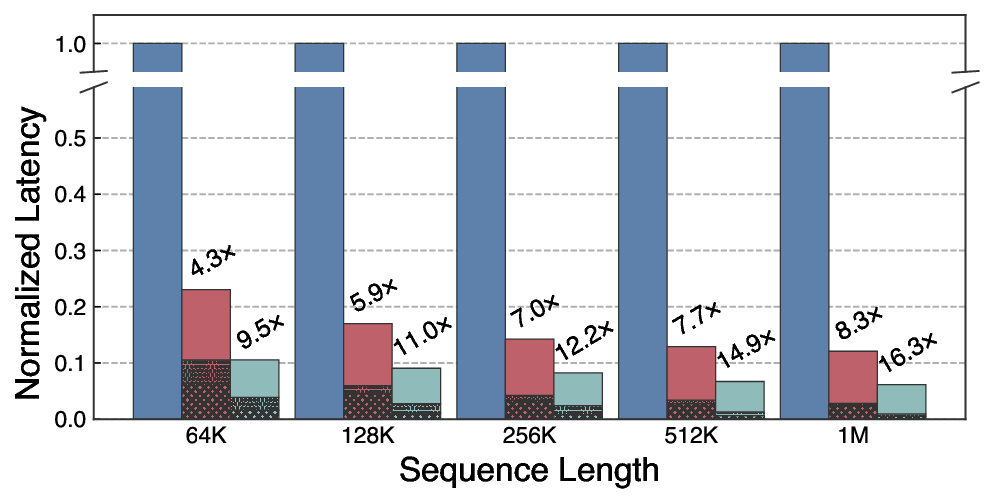}
    \vspace{-20pt}
    \caption*{(b) Batch size = 4}
  \end{minipage}\hfill
  \begin{minipage}[t]{0.33\linewidth}
    \centering
    \includegraphics[width=\linewidth]{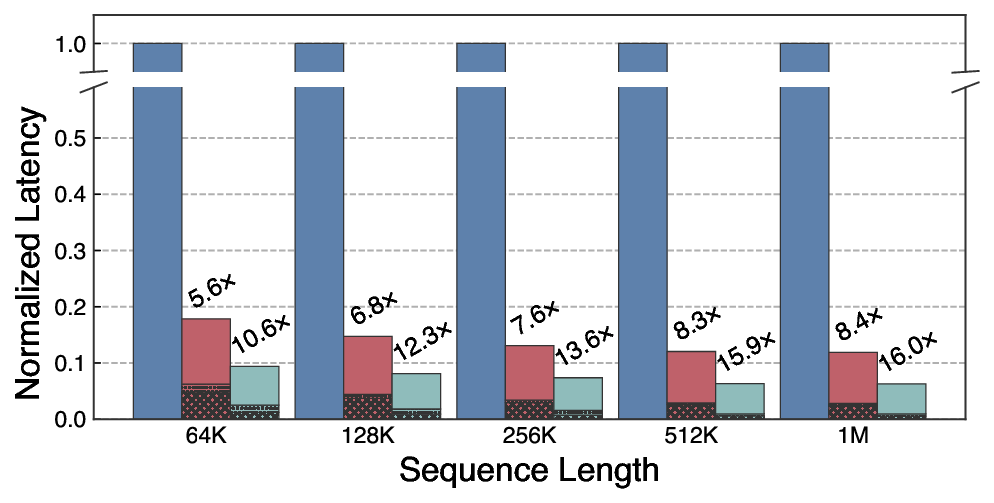}
    \vspace{-20pt}
    \caption*{(c) Batch size = 8}
  \end{minipage}
  \vspace{4pt}

  \begin{minipage}[t]{0.33\linewidth}
    \centering
    \includegraphics[width=\linewidth]{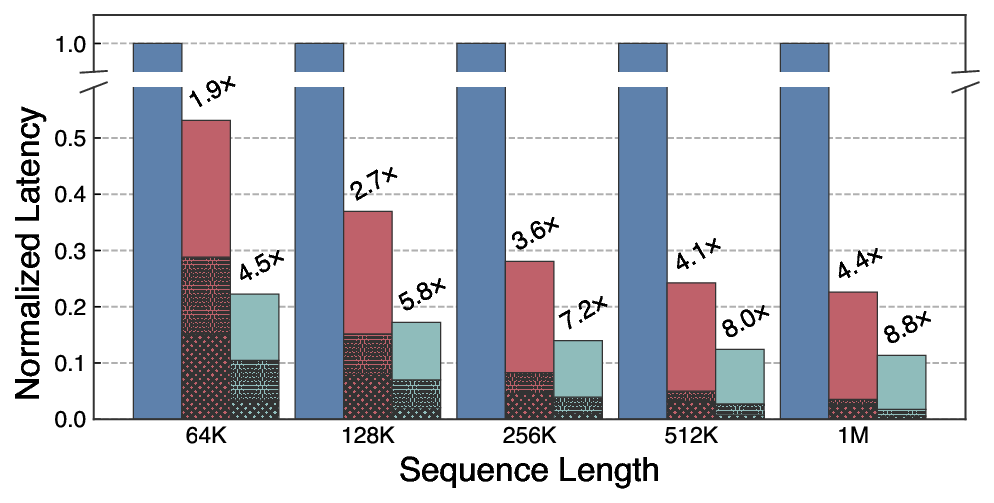}
    \vspace{-20pt}
    \caption*{(d) Batch size = 1}
  \end{minipage}\hfill
  \begin{minipage}[t]{0.33\linewidth}
    \centering
    \includegraphics[width=\linewidth]{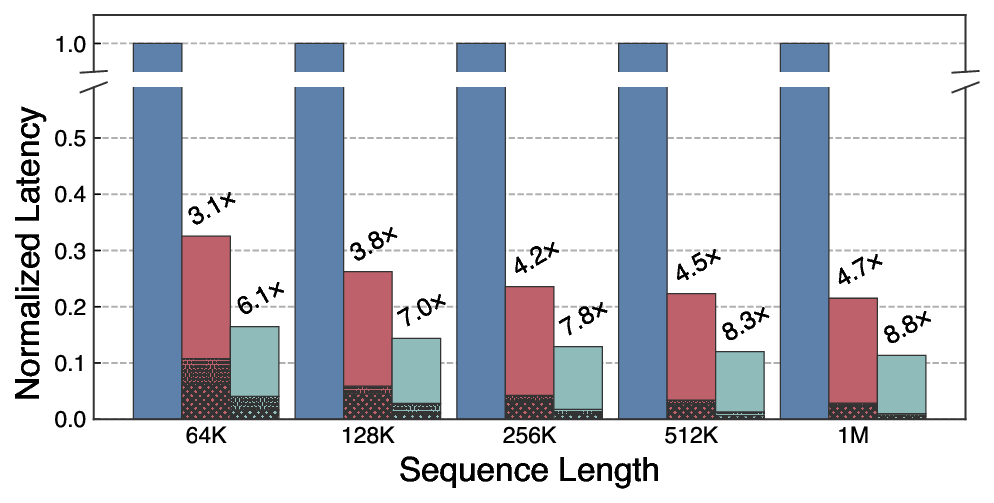}
    \vspace{-20pt}
    \caption*{(e) Batch size = 4}
  \end{minipage}\hfill
  \begin{minipage}[t]{0.33\linewidth}
    \centering
    \includegraphics[width=\linewidth]{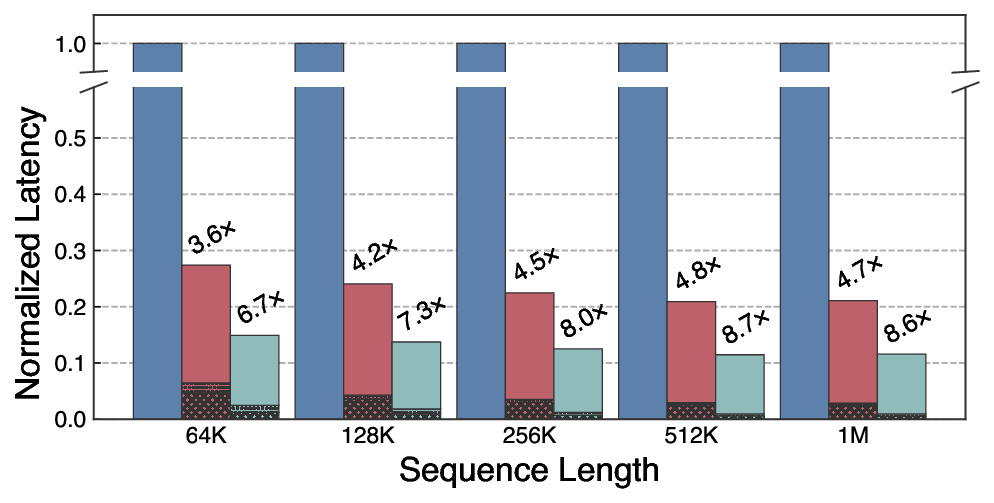}
    \vspace{-20pt}
    \caption*{(f) Batch size = 8}
  \end{minipage}

  \caption{Attention latency breakdown across batch sizes. Token selection rate: 6.25\% in (a)-(c) and 12.5\% in (d)-(f).}
  \label{fig:breakdown}
\end{figure*}

\textbf{Long Context Benchmarks.}
To answer \textbf{Q1}, we evaluate the accuracy of {\sys} on two widely-used long-context benchmarks: RULER~\cite{hsieh_ruler_2024} and LongBench~\cite{bai_longbench_2024}.
We compare {\sys} against the Full baseline (uncompressed) and Quest across three models: LLaMA-3.1-8B-Instruct, Qwen3-4B-Instruct, and Ministral-3-8B-Base.
The default compression configuration is used, consistent with the subsequent performance evaluation.

Table~\ref{tab:accuracy} shows the results.
For LongBench, {\sys} incurs only \textbf{0.86\%} average accuracy loss compared to the uncompressed baseline, and \textbf{0.64\%} compared to Quest.
On RULER, it is configured to test 64K-context task.
The average scores of {\sys} are 89.85, 90.47, 86.94 for LLaMA-3.1-8B, Qwen3-4B, and Ministral-3-8B respectively, incurring only \textbf{1.86\%}, \textbf{2.58\%}, and \textbf{3.55\%} accuracy loss compared to the uncompressed baseline.
Correspondingly, the accuracy loss compared to Quest are \textbf{0.55\%}, \textbf{1.74\%}, and \textbf{2.69\%}.

\begin{figure}[t]
  \centering
  \includegraphics[width=\linewidth]{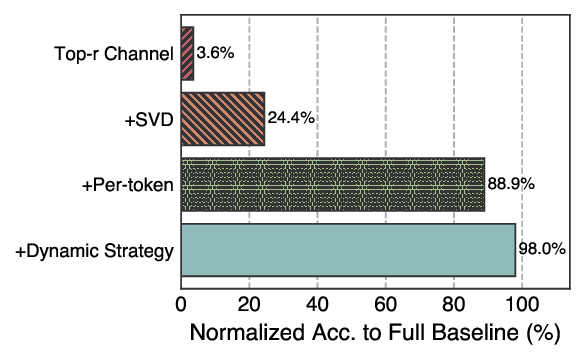}
  \caption{Accuracy with Different Optimizations.}
  \label{fig:ablation}
\end{figure}

\textbf{Accuracy with Different Optimizations.}
To further analyze how individual optimizations affect model accuracy, we present an incremental accuracy breakdown in Figure~\ref{fig:ablation}.
It shows the normalized accuracy relative to the uncompressed baseline on LongBench with LLaMA-3.1-8B.
Starting from a baseline that retains 10\% of tokens via sequence compression and applies uniform top-$r$ channel selection without SVD rotation, we incrementally add: (1) SVD rotation, (2) per-vector element selection, and (3) dynamic compression strategy.
SVD rotation and per-vector selection improve the normalized accuracy by 20.8\% and 64.5\%, respectively.
The dynamic compression strategy further improves the total accuracy to 98\%.


\begin{figure*}[t]
  \centering
  \begin{minipage}[t]{0.32\linewidth}
    \centering
    \includegraphics[width=\linewidth]{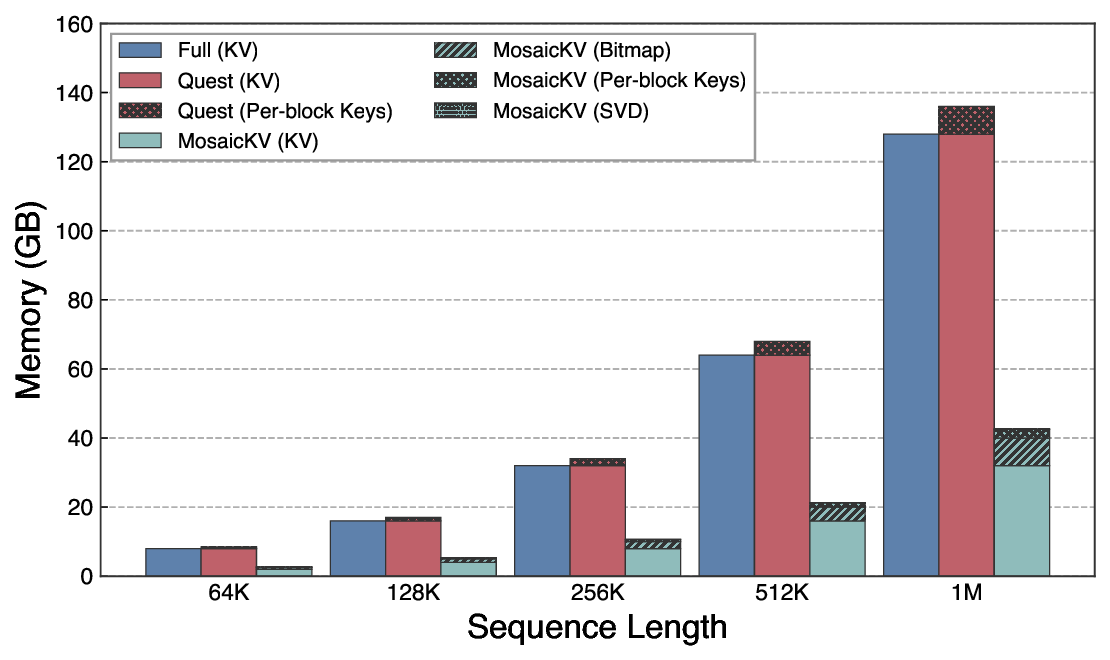}
    \vspace{-20pt}
    \caption*{(a) KV cache memory footprint}
  \end{minipage}\hfill
  \begin{minipage}[t]{0.32\linewidth}
    \centering
    \includegraphics[width=\linewidth]{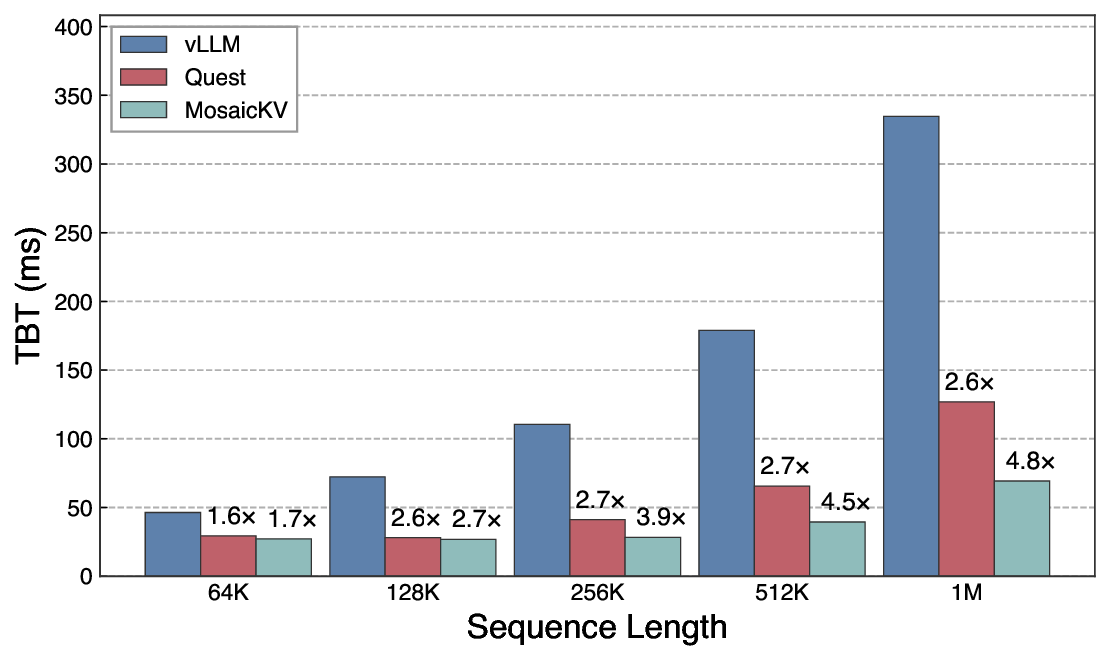}
    \vspace{-20pt}
    \caption*{(b) Decode Latency}
  \end{minipage}\hfill
  \begin{minipage}[t]{0.32\linewidth}
    \centering
    \includegraphics[width=\linewidth]{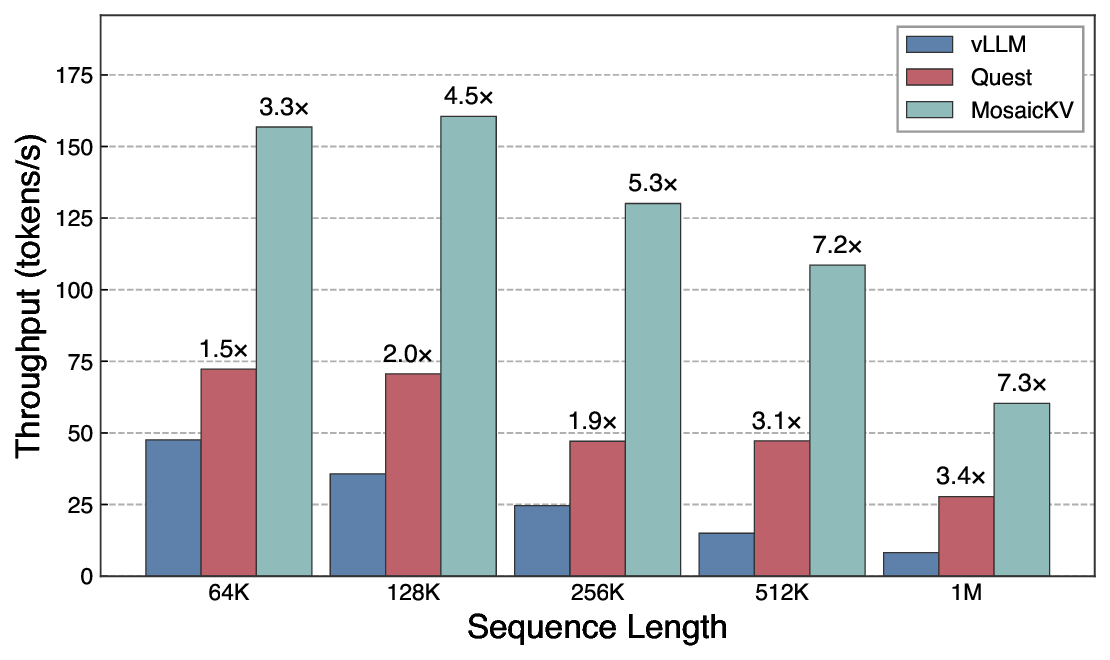}
    \vspace{-20pt}
    \caption*{(c) Decode throughput}
  \end{minipage}
  \caption{Performance of end-to-end long-context serving, with different context lengths.}
  \label{fig:e2e}
\end{figure*}

\subsection{Microbenchmark}
\label{sec:eval:micro}

\textbf{Memory Footprint.}
To answer \textbf{Q2}, we break down the KV cache memory footprint of {\sys} on LLaMA-3.1-8B across different context lengths, as shown in Figure~\ref{fig:e2e}(a).
The Full baseline stores the entire KV cache, while Quest additionally maintains per-block keys, adding 6.25\% memory overhead (block size 16).
With 25\% channel retention, {\sys} reduces the KV cache size by \textbf{3$\times$} and \textbf{3.18$\times$} compared to Full and Quest, respectively.

\textbf{Attention Latency.}
To answer \textbf{Q3}, we evaluate the attention latency of {\sys} on LLaMA-3.1-8B across different context lengths and batch sizes, as shown in Figure~\ref{fig:breakdown}.
For {\sys} and Quest, the latency is broken down into score computing (Score), token selection (TopK), and attention computation (Attn).
We further evaluate two token selection rates: 6.25\% in Figure~\ref{fig:breakdown}(a)-(c) and 12.5\% in Figure~\ref{fig:breakdown}(d)-(f).
{\sys} achieves up to \textbf{16$\times$} speedup in total attention latency over the Full baseline.
Compared to Quest, {\sys} achieves up to \textbf{2.4$\times$} speedup in total attention latency.
For score computing, {\sys} achieves an average of \textbf{4.2$\times$} speedup over Quest.




\subsection{End-to-End Serving Performance}
\label{sec:eval:e2e}

To answer \textbf{Q4}, we evaluate both the decode latency and throughput of {\sys} on LLaMA-3.1-8B across different context lengths.
We use vLLM~\cite{kwon_efficient_2023} as the Full baseline.
Due to the limited GPU memory of the H800, both vLLM and Quest cannot accommodate the entire KV cache for a 512K-token request, with batch size 1.
{\sys} extends this limit to about 1.44M tokens.
To evaluate longer contexts, we profile a subset of attention layers and extrapolate the total latency and throughput, as all transformer layers perform identical operations, for extremely long-context requests.

\textbf{Decode Latency.}
We evaluate the Time-Between-Tokens (TBT) as the decode latency metric, with results shown in Figure~\ref{fig:e2e}(b).
The batch size is 4 for all systems.
At 1M context length, {\sys} achieves \textbf{4.8$\times$} lower TBT than vLLM and \textbf{1.8$\times$} lower than Quest.
The average speedups are \textbf{3.8$\times$} and \textbf{1.5$\times$} over vLLM and Quest, respectively.

\textbf{Decode Throughput.}
Figure~\ref{fig:e2e}(c) shows the decode throughput of {\sys} and the two baselines.
Due to the reduced memory footprint, {\sys} supports a batch size 3$\times$ larger than vLLM and Quest.
At 1M context length, {\sys} delivers \textbf{7.3$\times$} higher throughput than vLLM and \textbf{2.2$\times$} higher than Quest.
The average speedups are \textbf{5.5$\times$} and \textbf{2.3$\times$} over vLLM and Quest, respectively.


\begin{table}[t]
  \centering
  \caption{Compression overhead to the prefill stage.}
  \label{tab:compression_overhead}
  \small
  \setlength{\aboverulesep}{2pt}
  \setlength{\belowrulesep}{2pt}
  \renewcommand{\arraystretch}{0.85}
  \begin{tabular*}{\columnwidth}{l@{\extracolsep{\fill}}rrrrr}
    \toprule
    & \textbf{64K} & \textbf{128K} & \textbf{256K} & \textbf{512K} & \textbf{1M} \\
    \midrule
    Prefill (s) & 13.7 & 57.3 & 241.9 & 1001.5 & 4093.9 \\
    Overhead (s) & 1.0 & 2.1 & 4.1 & 8.3 & 16.6 \\
    Percentage (\%) & 7.6\% & 3.6\% & 1.7\% & 0.8\% & 0.4\% \\
    \bottomrule
  \end{tabular*}
\end{table}

\textbf{Compression Overhead.}
{\sys} introduces additional operations, including SVD computation and compression, to the prefill stage.
We evaluate the total overhead during prefill across different input token lengths.
The results are shown in Table~\ref{tab:compression_overhead}.
The total overhead is 7.6\% at 64K tokens and decreases to 0.4\% at 1M tokens.
During the decode stage, compressing the entire update buffer costs only 40.6\,$\mu$s, which is negligible relative to the attention computation.

\section{Discussion}
\label{sec:discuss}


\textbf{Compatibility.}
{\sys} is compatible with diverse model architectures.
It is applicable to all attention-based LLMs, including those employing Grouped-Query Attention (GQA)~\cite{ainslie-etal-2023-gqa} and Multi-head Latent Attention (MLA)~\cite{deepseekai2024deepseekv2strongeconomicalefficient}.
Although these architectures share KV cache across query heads, {\sys} can still reduce the KV cache size.
Notably, the LLaMA-3.1-8B model used in our evaluation adopts GQA.
Furthermore, Mixture of Experts (MoE) models~\cite{shazeer2017outrageouslylargeneuralnetworks, fedus2022switchtransformersscalingtrillion} also incorporate attention layers and can therefore benefit from {\sys}.

{\sys} is also compatible with commonly-used LLM serving optimizations.
It can be integrated with PagedAttention~\cite{kwon_efficient_2023} and RadixAttention~\cite{sglang}, as the full token sequence is retained and page-level scheduling and prefix sharing remain unaffected.
Moreover, {\sys} is compatible with FlashAttention~\cite{dao2022flashattention}, upon which our implementation is built (Section~\ref{sec:system:kernels}).

\textbf{KV Cache Offloading.}
{\sys} stores the entire KV cache in GPU memory.
Existing long-context approaches~\cite{xiao_infllm_2024, liu_retrievalattention_2025, chen2024magicpiglshsamplingefficient} offload the KV cache to CPU memory and selectively fetch the required KV vectors to the GPU for attention computation.
However, such methods introduce additional PCIe traffic on the critical decode path, which becomes a bottleneck for decode latency and throughput.
Several optimizations have been proposed to mitigate this issue, including GPU-side KV cache caching and accelerated KV cache fetching~\cite{xiao_infllm_2024,chen2025retroinfervectorstorageapproachscalable,hao2025omnikv}.
These offloading methods and their associated optimizations are complementary to {\sys}, which can still achieve a high compression rate and corresponding performance improvement.
Furthermore, since {\sys} compresses both K and V vectors, it can further reduce the PCIe traffic by fetching only the compressed vectors to the GPU.

\textbf{Combining with Chunked Prefill.}
First, {\sys} can still optimize the decode stage when chunked prefill~\cite{amey_chunk_prefill_2024} is employed.
Second, with chunked prefill, {\sys} can further compress the KV vectors of input tokens to accelerate the prefill stage.
Chunked prefill generates the KV cache for input tokens at chunk granularity.
Analogous to compressing output tokens during the decode stage, {\sys} can compress these input chunks' KV vectors using the heterogeneous double compression buffering.

\textbf{Combining with Quantization.}
{\sys} is orthogonal to quantization techniques and can be combined with them.
Quantizing the elements of the compressed KV cache can directly reduce the memory footprint.
Moreover, existing works such as DiffKV~\cite{diffkv} propose importance-aware token quantization.
By combining with such approaches, {\sys} can assess the importance of each compressed element and apply appropriate quantization accordingly.

\textbf{Limitations and Future Work.}
First, {\sys} currently focuses on accelerating the decode stage of long-context requests, which is particularly relevant as thinking models become increasingly prevalent.
As discussed previously, {\sys} can further accelerate the prefill stage when combined with chunked prefill, which represents a promising direction for future work.
Second, {\sys} currently leverages only CUDA cores for PackedAttention.
This is sufficient at present, as memory bandwidth remains the bottleneck even on server-grade HBM-equipped GPUs.
However, if future HBM memory achieves substantially higher bandwidth such that memory is no longer the bottleneck, CUDA cores may limit {\sys}'s performance.
A potential solution is to combine both tensor cores and CUDA cores to accelerate PackedAttention.

\section{Related Work}
\label{sec:related}

\textbf{KV Cache Compression.}
Existing works compress the KV cache by selecting important tokens~\cite{beltagy_longformer_2020, zaheer_big_2020, lu_moba_2025, yuan_native_2025, zhang_h2o_2023, liu_scissorhands_2023, li_snapkv_2024,chen2024magicpiglshsamplingefficient,xiao2024duoattentionefficientlongcontextllm,xiao2024efficientstreaminglanguagemodels,lee_infinigen_2024,tang_quest_2024} or important channels~\cite{ribar_sparq_2024,xu_think_2024,liao2025sparkqueryawareunstructuredsparsity,mustafar2025,sun_shadowkv_2025, chang_palu_2024}.
Quest~\cite{tang_quest_2024} performs query-aware token selection.
Approaches such as MoBA~\cite{lu_moba_2025} and NSA~\cite{yuan_native_2025} retrain the model to achieve higher accuracy under sequence compression.
InfiniGen~\cite{lee_infinigen_2024} accelerates token selection by compressing the channels of key vectors.
However, they cannot perform channel compression on the KV cache during attention computation.
SparK~\cite{liao2025sparkqueryawareunstructuredsparsity} applies fine-grained unstructured compression to K but requires decompression before each attention step.
Unlike existing systems, {\sys} performs per-token dynamic two-dimensional compression throughout the entire decode stage, including both token selection and attention computation, achieving high memory compression rates, high performance, and low accuracy loss.


\textbf{Quantization.}
Quantization methods~\cite{hooper_kvquant_2024,liu_kivi_2024, zhao2024atomlowbitquantizationefficient, kang2024gearefficientkvcache, yankun2025svdq125bit410xkey} are also employed to reduce the KV cache size.
KVQuant~\cite{hooper_kvquant_2024} observes that element-level outliers are the key challenge for low-bit KV cache quantization and proposes per-vector dense-and-sparse quantization to isolate outliers.
DiffKV~\cite{diffkv} applies importance-aware KV cache quantization.
As discussed previously, {\sys} can be combined with quantization techniques to achieve further memory reduction.

\textbf{Sparse GPU Kernels.}
Traditional sparse matrix kernels, such as cuSPARSE~\cite{cusparse}, are primarily designed for extreme sparsity ($>$99\%).
They record the location of each non-zero element to reduce memory consumption.
NVIDIA provides a sparse tensor operation~\cite{mishra_accelerating_2021} that requires 2:4 structured sparsity.
PIT~\cite{zheng_pit_2023} densifies sparse tensors via permutation invariant transformation.
FlashInfer~\cite{ye2025flashinfer} accelerates block-sparse attention.
{\sys} leverages the underutilized CUDA cores and the packed KV format to efficiently execute the decode stage with sparse KV cache.

\section{Conclusion}
\label{sec:concl}

This paper presented {\sys}, a long-context LLM inference framework that combines channel and sequence KV cache compression to overcome the memory bottleneck imposed by long context requests.
To address the accuracy challenge of two-dimensional compression, {\sys} introduces dynamic two-D compression, which exploits the non-uniform importance distribution of the KV cache through per-vector element selection and a dynamic compression strategy that partitions the KV cache into segments with tailored strategies.
To translate the high compression rate into real performance gains, {\sys} introduces compressed KV cache management, which leverages the underutilized CUDA cores and CPU during decode-stage attention to perform packed sparse attention over the compressed KV cache and heterogeneous double compression buffering for asynchronous KV cache management.
Evaluation on an H800 GPU across multiple models and benchmarks shows that {\sys} achieves up to $16\times$ attention speedup, $4.8\times$ lower decode latency, and $7.3\times$ higher throughput, while reducing KV cache memory by $3\times$ with only 1.76\% average accuracy loss.

\bibliographystyle{ACM-Reference-Format}
\bibliography{references}

\end{document}